\documentclass[conference]{IEEEtran}
\IEEEoverridecommandlockouts
\usepackage{cite}
\usepackage{amsmath,amssymb,amsfonts}
\usepackage{algorithmic}
\usepackage{graphicx}
\usepackage{textcomp}
\usepackage{xcolor}
\usepackage{booktabs}
\usepackage{soul}
\usepackage{subcaption}
\def\BibTeX{{\rm B\kern-.05em{\sc i\kern-.025em b}\kern-.08em
    T\kern-.1667em\lower.7ex\hbox{E}\kern-.125emX}}
\begin{document}

\title{Variational Neural Networks}

\author{\IEEEauthorblockN{Illia Oleksiienko$^*$, Dat Thanh Tran$^\dag$ and Alexandros Iosifidis$^*$, \IEEEmembership{Senior Member, IEEE}
\thanks{This work has received funding from the European Union’s Horizon 2020 research and innovation programme (grant agreement No 871449 (OpenDR)).}
\thanks{I. Oleksiienko and A. Iosifidis are with the Department of Electrical and Computer Engineering, Aarhus University, Denmark (e-mail: \{io,ai\}@ece.au.dk). D. T. Tran is with the Department of Computing Sciences, Tampere University, Finland (e-mail: thanh.tran@tuni.fi).} }
}

\maketitle

\begin{abstract}
Bayesian Neural Networks (BNNs) provide a tool to estimate the uncertainty of a neural network by considering a distribution over weights and sampling different models for each input. In this paper, we propose a method for uncertainty estimation in neural networks which, instead of considering a distribution over weights, samples outputs of each layer from a corresponding Gaussian distribution, parametrized by the predictions of mean and variance sub-layers. In uncertainty quality estimation experiments, we show that the proposed method achieves better uncertainty quality than other single-bin Bayesian Model Averaging methods, such as Monte Carlo Dropout or Bayes By Backpropagation methods.
\end{abstract}


\section{Introduction}
The ability to estimate the uncertainty of prediction in neural networks provides advantages in using high-performing models in real-world problems, as it enables higher-level decision-making to consider such information in further actions. To do so, one needs the neural network to accompany its output with a measurement of its corresponding uncertainty for each input it processes. Several approaches have been introduced to this end, with Bayesian Neural Networks (BNNs) \cite{1995bayessian_nn, wilson2020bayesian, charnock2020bayesian} providing an elegant framework for estimating uncertainty of a neural network by introducing a probability distribution over its weights and sampling different models that are meant to describe the input from different points of view. This allows to determine inputs for which the network predictions are different, leading to a measurement of the network uncertainty in its outputs. Such an approach usually comes with an increased computational cost, but may be valuable for tasks where prediction errors result in high losses.

The choice of the weights prior probability distribution function influences the statistical quality of the model and the computational resources needed to use such neural networks. This creates a possibility to explore different approaches to BNNs by using Gaussian \cite{blundell2015weight}, Bernoulli \cite{2016dropout}, Categorical \cite{osband2018randomized} or other distributions. 
Sampling from the posterior distribution can be difficult, due to the complex nature of it.
This leads to methods that avoid direct computation of the posterior, such as Markov Chain Monte Carlo (MCMC) \cite{hastings1970mcmc} which constructs a Markov chain of samples $S_i$ that are distributed following the desired posterior, or Variational Inference \cite{Blei2017vi} which scales better than MCMC and aims to estimate a parametrized distribution that should be close to the exact posterior.
How close the distributions are is computed using the Kullback-Leibler (KL) divergence  \cite{1951_kl_information}, but it still requires the exact posterior. This is overcome by computing an Evidence Lower Bound (ELBO) instead and optimizing it with Stochastic Variational Inference (SVI) \cite{hoffman2012svi}.

\begin{figure}
     \centering
     \begin{subfigure}[b]{0.55\linewidth}
         \centering
         \includegraphics[width=0.9\linewidth]{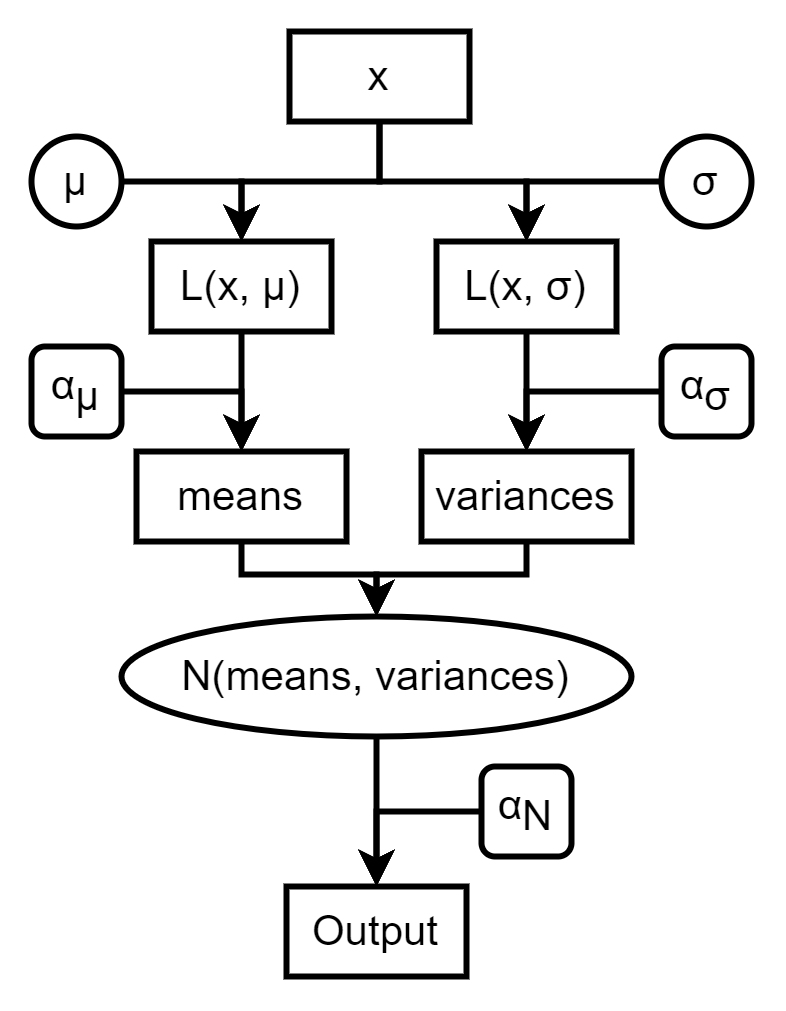}
         \caption{Computational graph of a layer in the proposed VNNs.}
         \label{fig:comp-graph-vnn}
     \end{subfigure} \:\:
     \begin{subfigure}[b]{0.39\linewidth}
         \centering
         \includegraphics[width=0.9\linewidth]{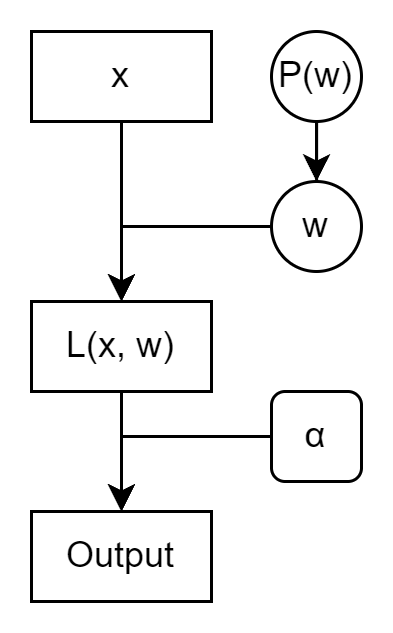}
         \caption{Computational graph of a layer in BNNs.}
         \label{fig:comp-graph-bnn}
     \end{subfigure}
    \caption{Comparison of computational graphs of (a) the proposed VNNs, and (b) BNNs. BNNs consider a distribution $P(w)$ over weights and sample different weights during each inference. VNNs consider a constant set of weights and use them to generate parameters of a Gaussian distribution for each layer, outputs of which are sampled from the corresponding distribution. Layer weights are represented by $\mu, \sigma, w$, activation functions by $\alpha, \alpha_{\mu}, \alpha_{\sigma}, \alpha_{N}$, classical layers by $L(\cdot)$, and $N(\cdot)$ represents the Gaussian distribution.}
    \label{fig:comp-graphs}
\end{figure}
We introduce Variational Neural Networks (VNNs) which do not consider a distribution over weights, but define sub-layers to generate parameters for the output distribution of the layer.
To keep computational and memory resource usage practical, we consider a Gaussian distribution with learnable mean and variance.
This is achieved by using two instances of the same regular layer like convolutional or linear with different weights, and using their predictions from the inputs as means and variances of the Gaussian distribution over the outputs.
We provide a neural network formulation that describes both related BNNs and the proposed VNNs in a unified manner, and show that VNNs, while being in the same group as Monte Carlo Dropout (MCD) \cite{2016dropout} and Bayes By Backprop (BBB) \cite{blundell2015weight} from the Bayesian Model Averaging (BMA) perspective \cite{wilson2020bayesian}, achieve better uncertainty quality and retain it with the increasing data dimensionality.
Fig. \ref{fig:comp-graphs} shows the difference between the computational graphs of the proposed VNNs (Fig. \ref{fig:comp-graph-vnn}) and the BNNs (Fig. \ref{fig:comp-graph-bnn}).

\section{Bayesian Neural Networks} \label{sec:bnn}
BNNs \cite{1995bayessian_nn, wilson2020bayesian, charnock2020bayesian,magris2022bayes_survey} consider a distribution over their weights $p(w|a)$ and a distribution over their hyperparameters $p(a)$. A predictive distribution over an output $y$ for a data point $x$ can be obtained by integrating over all possible hyperparameters and model weights, i.e.:
\begin{equation}
    p(y|x) = \int \int p(y|x, w) \: p(w|a) \: p(a) \: da \: dw.
\end{equation}
Given a dataset $D=(X_t, Y_t)$, where $X_t$ and $Y_t$ are the sets of inputs $\{x\}$ and targets $\{y\}$, the distribution of weights can be derived from Bayes' theorem as $p(w|D, a) = \frac{p(Y_t|X_t, w) p(w|a)}{p(D)}$, and the corresponding predictive distribution has a form
\begin{equation}
    p(y|x, D) = \int \int p(y|x, w) \: p(w|D, a) \: p(a|D) \: da \: dw. \label{eq:predictive_distribution}
\end{equation}

Classical neural networks can be viewed as BNNs with $p(a|D) = \delta(a-\hat{a})$ and $p(w|D, a) = \delta(w-\hat{w}_a)$ \cite{charnock2020bayesian}, where $\hat{a}$ are the selected model hyperparameters, $\hat{w_a}$ are the weights, optimized by training the model, and $\delta(x)$ is the Dirac delta function which has values $0$ everywhere except at $x=0$ where it equals to 1. In this case, the predictive distribution becomes
\begin{align}
\begin{split}
    p(y|x, D) &= \int \int p(y|x, w) \: p(w|D, a) \: p(a|D) \: da \: dw \\
    &= \int \int p(y|x, w) \: \delta(w-\hat{w}_a) \: \delta(a-\hat{a}) \: da \: dw \\
    &= p(y|x, \hat{w}_{\hat{a}}), \\
\end{split}
\end{align}
which is a distribution dictated by a loss of the network.

When training classical neural networks, hyperparameters are considered fixed at point $\hat{a}$ and weights are optimized either by maximum likelihood estimation (MLE), i.e.:
\begin{align}
\begin{split}
    w_{\mathrm{mle}} &= \underset{w}{\mathrm{argmax}} \Big[\log \, p(D|w, \hat{a}) \Big] \\
            &= \underset{w}{\mathrm{argmax}} \left[\sum_i \log \, p(Y_i|X_i, w, \hat{a})\right],
\end{split}
\end{align}
or by maximum a posteriori (MAP), i.e.:
\begin{align}
\begin{split}
    w_{\mathrm{map}} &= \underset{w}{\mathrm{argmax}} \Big[\log \, p(w|D, \hat{a})\Big] \\
            &= \underset{w}{\mathrm{argmax}} \Big[\log \, p(D|w, \hat{a}) + \log \, p(w)\Big],
\end{split}
\end{align}
where $\log \, p(w)$ is a regularization term.

Due to the complexity of neural networks, direct computation of $w_{\mathrm{mle}}$ or $w_{\mathrm{map}}$ cannot be achieved, and therefore approximate methods are used to find these values.
The most popular process to estimate the weight values is the Backpropagation algorithm \cite{kelley1960backprop}, where an initial randomly selected $w$ is updated following the direction of negative gradient of the loss function with respect to $w$.

\section{Related Works}\label{sec:related_works}

The use of BNNs in real-world applications is limited due to the complex nature of the possible prior and predictive distributions. Therefore, simplified versions are used. 
Assumptions that are proposed in different methods below aim to reduce memory, inference and training time, but they come with the cost of reducing the statistical quality of the resulting models. 
This problem is further discussed in Section \ref{sec:uncertainty_experiments}.

MCD \cite{2016dropout} considers a neural network with Dropout \cite{srivastava2014dropout} added to each layer. The Dropout layer effectively turns off random neurons of the layer by multiplying connection weights with a random binary mask sampled from a Bernoulli distribution. 
This allows to avoid overfitting specific neurons. 
After training, standard neural networks replace Dropout with a scaled identity function and all neurons are used for inference. 
Instead of replacing Dropout with identity, MCD uses it during inference leading to a stochastic model. The model uncertainty for an input is computed by performing inference multiple times and computing mean and variance of predictions. 
BBB \cite{blundell2015weight} samples model parameters from a Gaussian distribution and trains it using regular Backpropagation.
By doing so, the family of models with different weights is sampled from the learned distribution, and the uncertainty of the network is computed as the variation in predictions of different samples. 

Ensembles of neural networks \cite{osband2018randomized} can also be used for uncertainty estimation.
Ensembles are trained in parallel for the same task, but with different random seeds, which results in different weight initialization and training order.
Outputs from members of an ensemble will vary, and this can be used to improve performance by taking an average of their predictions, or to estimate uncertainty by computing the variance of their outputs. Such an approach can be viewed as a BNN with a categorical distribution over weights that randomly selects one of the trained model weights. 
Hypermodels \cite{dwaracherla2020hypermodels} use an additional model $\theta = g_{\nu}(\mathbf{z})$ to generate parameters for a base model $f_{\theta}(x)$.
Linear Hypermodels set $g_{\nu}(\mathbf{z}) = a + B\mathbf{z}, \:\:\mathbf{z} \sim \mathcal{N}(0, I)$. Using different samples of $\mathbf{z}$, one can sample different model parameters and estimate uncertainty in the same way as for the aforementioned methods.

\section{Variational Neural Networks} \label{sec:vnn}
As introduced in Section \ref{sec:bnn}, a neural network is described by its weights $w$ and hyperparameters $a$. Hyperparameters include the structure of the network, i.e., the type and number of layers, their size and connections.
Usually, we limit the hyperparameters by defining some of them in the beginning, e.g., by selecting that we want to use convolutional layers.
This is a reasonable approach, as it is impractical to iterate through all possible types and structures of networks during training.
We are using the neural network formulation $\mathrm{NN}(x) := F^{\Lambda}(x, w)$, 
where $\mathrm{NN}(x)$ is a neural network applied to an input $x$, $w$ are the trained weights, $F$ is a neural network function that incorporates structure and other hyperparameters inside it, and $\Lambda$ is a set of layer implementations, which are used by $F$ to process layer inputs.

In case of CNNs, $\Lambda=\{\mathrm{Conv2D}(x, w), \mathrm{FC}(x, w)\}$ results in a regular CNN, where $\mathrm{Conv2D}(x, w)$ is a 2D convolutional layer function and $\mathrm{FC}(x, w)$ is a fully connected layer function.
If we select $\Lambda=\{\mathrm{Conv2D}(x, w_c \sim \mathcal{N}(\mu_c, \Sigma_c)), \mathrm{FC}(x, w_l \sim \mathcal{N}(\mu_l, \Sigma_l))\}$ with layer weights sampled from a corresponding Gaussian distribution, then the resulting network is a BBB CNN.
Such neural network formulation allows to accurately describe all the discussed uncertainty estimation methods, as well as the proposed VNNs.

\subsection{Variational Layer}\label{sec:vl}
We define a Variational Layer (VL) that takes an input $x$ and weights $w$ as
\begin{align}
\begin{split}
    &\mathrm{VL}(x, w) = \alpha_{\mathcal{N}}(f(x, w)), \\
    &f(x, w) \sim \mathcal{N}\Big(\alpha_\mu(L(x, \mu)), diag[(\alpha_\sigma(L(x, \sigma)))^2]\Big), \\
    &w = (\mu, \sigma),
    \label{eq:vl}
\end{split}
\end{align}
where $L(x, w)$ is a regular neural network layer, such as fully connected, convolutional or a recurrent layer. $L(x, \mu)$ and $L(x, \sigma)$ represent instances of the same layer with different values of parameters and corresponding activation functions $\alpha_\mu(\cdot)$ and $\alpha_\sigma(\cdot)$. The activation function $\alpha_{\mathcal{N}}(\cdot)$ can be used to apply nonlinearity to the randomly sampled values $f(x, w)$. By selecting which of $\alpha_\mu(\cdot)$, $\alpha_\sigma(\cdot)$, $\alpha_{\mathcal{N}}(\cdot)$ are set to identity and which are set to actual activation functions (such as the Rectified Linear Unit (ReLU)) one can create networks that are described by different mathematical models. In the following, we show how different selections can lead to specific types of uncertainties, i.e., epistemic and aleatoric uncertainties \cite{kendal2017uncertainties}.

Training of VNNs is performed by averaging outputs from different network passes for the same sample and, thanks to the reparametrization trick \cite{kingma2014autoencoding}, the regular Backpropagation algorithm is applied.
Networks can also be trained with a single pass, which results in the same training procedure as for classical neural networks. The models are trained with the usual loss functions that are suitable for the task.

\subsection{Output uncertainty estimation} \label{sec:output_uncertainty}

Estimation of prediction uncertainties in VNNs and BNNs can be done following the same formulation, but it obtained from different characteristics of these methods.
Below, we first describe how BNNs can be reformulated by splitting the parametrized distribution over weights into isolated parameters and a non-parametric distribution, and then show that this formulation can be applied to VNNs.

Following \cite{osband2021epistemic}, we consider a neural network $F(x, w)$ with a parametric distribution over weights $q_m(w)$.
We assume the choice of $q_m(w)$ in a form
\begin{equation}
    w = Q(m, z), \:\:\:\:\:\:\:\: w \sim q_m(w), \:\:\:\:\:\:\:\: z \sim p(z),
\end{equation}
where $p(z)$ is a non-parametric distribution and $Q(\cdot)$ applies a deterministic transformation, parametrized by $m$, to a non-parametric random variable $z$.
Such formulation is suitable for every uncertainty estimation method described in Section \ref{sec:related_works}.
For BBB models, $Q(\cdot)$ is defined as
\begin{equation}
    Q(m, z)  = \mu + \sigma^2 z, \:\:\:\:\:\:\:\: z \sim \mathcal{N}(0, I), \:\:\:\:\:\:\:\: m = (\mu, \sigma),
\end{equation}
where we break down a parametric Gaussian distribution $\mathcal{N}(\mu, \sigma I)$ into two parts: a parametric deterministic transformation $z \longrightarrow \mu + \sigma^2 z$ and a non-parametric random variable $z \sim \mathcal{N}(0, I)$.

Defining an epistemic index $z \sim p(z)$ \cite{osband2021epistemic}, we can formulate a deterministic neural network $F_d(\cdot)$ function that takes a draw of a random non-parametric variable $z$, instead of using $F(\cdot)$ with a complex distribution over $w$:
\begin{equation}
    F_d(x, m, z) := F(x, w), \:\:\:\:\:\:\:\: w = Q(m, z), \:\:\:\:\:\:\:\: z \sim p(z).
\end{equation}
With this formulation, a predictive distribution (\ref{eq:predictive_distribution}) for fixed hyperparameters is defined by splitting $w$ into $m$ and $z$ as follows \cite{osband2021epistemic, wilson2020bayesian}:
\begin{small}
\begin{align}
\begin{split}
    p(y|x, D) &= \int p(y|x, w) \: q_m(w|D) \: dw = \int p(y|x, m, z) \: p(z) \: dz, \\
    \mathbb{E}[y] &= \int y \, p(y|x, D) dy  \approx \frac{1}{T} \sum^T_{i} F_d(x, m, z_i), \\
    \mathrm{Cov}[y] &= \int (y - E[y])(y - E[y])^T p(y|x, D) \: dy, \\
    &\approx \frac{1}{T} \sum^T_{i} (E[y] - F_d(x, m, z_i))(E[y] - F_d(x, m, z_i))^T, \label{eq:predictive_indexed}
\end{split}
\end{align}
\end{small}
where expectation and variance are computed using Monte Carlo integration, which can be viewed as an approximation of $p(z)$ with $\sum^T_{i=0} \frac{\delta(z - z_i)}{T}, \:\:z_i \sim p(z), \:\:i \in {1,\dots, T}$ \cite{wilson2020bayesian}.
Variance of the outputs is computed by taking main diagonal values of the $\mathrm{Cov}[y]$ representing the uncertainty of the model.

VNNs, despite not having a direct distribution over weights, can also be formulated as a deterministic function $F_d(x, w, z)$ with a variational index $z \sim p(z)$.
This is done by describing the output Gaussian distribution $\mathcal{N}\Big(\alpha_\mu(L(x, \mu)), diag[(\alpha_\sigma(L(x, \sigma)))^2]\Big)$ of a VL as a linear transformation of a unit Gaussian $\alpha_\mu(L(x, \mu)) + diag[(\alpha_\sigma(L(x, \sigma)))^2] \mathcal{N}(0, I)$.

\subsection{Epistemic uncertainty} \label{sec:vnn_epistemic_uncertainty}
Epistemic uncertainty describes the lack of knowledge of the model and can be improved by providing a better model structure, better dataset or improved training procedure, while aleatoric uncertainty describes the uncertainty in data due to noise in data perceiving process or domain shift \cite{hullermeier2021epiale, gawlikowski2021uncertaintyindl}. 
Usually, the epistemic uncertainty in BNNs is modeled by assuming a distribution over weights and fixed hyperparameters.
The use of unfixed hyperparameters leads to the Hierarchical Bayes approach \cite{allenby2006hierarchicalbayes}, where the epistemic uncertainty is represented by both hyperparameters and weight distributions.

Given the fact that model parameters and structure are usually separated, the predictive distribution equation (\ref{eq:predictive_distribution}) holds only in the case where the hyperparameters' influence is limited to the training procedure.
If the model structure is included in hyperparameters, then:
\begin{equation}
    p(y|x, D) = \int \int p(y|x, w, a) \: p(w|D, a) \: p(a) \: da \: dw, \label{eq:correct_predictive_distribution}
\end{equation}
where the probability of a prediction for a selected model, depends on both weights and hyperparameters.
Following this approach, the predictive distribution of VNN (\ref{eq:predictive_indexed}) can be interpreted as a predictive distribution, computed for a Hierarchical BNN with a unit Gaussian distribution over hyperparameters $z$ and a Dirac delta distribution over weights:
\begin{align}
\begin{split}
    p(y|x, D) &= \int \int p(y|x, w, z) \: p(w|D, z) \: p(z) \: dz \: dw \\
    &= \int \int p(y|x, w, z) \: \delta(w-\hat{w}) \: p(z) \: dz \: dw \\
    &= \int p(y|x, \hat{w}, z) \: p(z) \: dz. \\
\end{split}
\end{align}
This formulation shows that the use of the variational index $z$ models the epistemic uncertainty in VNNs.

\subsection{Aleatoric uncertainty}\label{sec:vnn_aleatoric_uncertainty}
Fully connected and convolutional layers can be described as the operation $L(x, \lambda) = W_\lambda x + b_\lambda$, where $\lambda=(W_\lambda, b_\lambda)$, $W_\lambda$ and $b_\lambda$ are weights and biases of the layer, and a corresponding activation function $\alpha_\lambda(\cdot)$ can be applied to the output of $L(x, \lambda)$.

Consider a Variational Layer with $L(x, \sigma)=\Sigma, \:\:\Sigma \in \mathbb{R}$, which can be directly achieved by setting $W_\sigma=0, \:\:b_\sigma = \Sigma$ and setting the corresponding activation function $\alpha_\sigma(\cdot)$ to identity.
Applying the formulation of fully connected and convolutional layers to $f(x, w)$ (\ref{eq:vl}) and using the reparametrization trick \cite{kingma2014autoencoding}, we can reformulate it as follows:
\begin{align}
\begin{split}
    \epsilon & \sim \mathcal{N}(0, I), \\
    f(x, w) &= \alpha_\mu(W_\mu x + b_\mu) + \Sigma \epsilon = L(x, \mu) + \epsilon_\sigma, \\
    \epsilon_\sigma & \sim \mathcal{N}(0, \Sigma I). \label{eq:reparametrization_aleatoric}
\end{split}
\end{align}
In this formulation, $\epsilon_\sigma$ models the aleatoric uncertainty \cite{hullermeier2021epiale} for the next subnetwork, which takes outputs of the current layer as inputs and cannot improve this uncertainty by improving the model.

\section{Experiments} \label{sec:uncertainty_experiments}
A recently proposed framework called Epistemic Neural Networks \cite{osband2021epistemic} aims to provide a possibility to rank BNNs based on their ability to accurately estimate output uncertainty.
This is done by first generating a synthetic dataset $D_T = \{(x, y)_t$ for $t \in [0, T-1]\}$ for a simple regression task $y = f(x) + \epsilon$, where $y$ is an output scalar, $x$ is an input data point with $D_x$ number of dimensions, $\epsilon$ is a random variable sampled from a Gaussian distribution $\mathcal{N}(0, {\sigma}^2)$ representing an aleatoric uncertainty.
The dataset size $T$ is determined as $T = D_x \lambda$, where $\lambda$ is a hyperparameter, meaning that more data points are created for a higher dimensionality of $x$.
The dataset is used to train a Neural Network Gaussian Process (NNGP) \cite{lee2018nngp} and an uncertainty estimation model of interest.
NNGP serves as an ideal probabilistic model for this data, and a predictive distribution of a selected uncertainty estimation model should be as close as possible to the predictive distribution of the NNGP model. 
The above process is used to create two datasets, one used for training the uncertainty estimation model and one (test set) used to evaluate the uncertainty estimation performance. Following \cite{osband2021epistemic}, random noise is added to the data belonging to the training set, as it has been shown to increase the uncertainty estimation performance, which is measured by computing the KL-divergence between the true posterior $\mathcal{N}(\mu_{\mathrm{GP}}, k_{\mathrm{GP}})$ and a model predictive distribution $\mathcal{N}(\mu_{B}, k_{B})$. Lower values of KL-divergence represent better uncertainty quality for a selected model, and therefore can be used to rank different approaches for uncertainty estimation.

\begin{figure}
     \centering
     \begin{subfigure}[b]{1\linewidth}
         \centering
         \includegraphics[width=0.9\linewidth]{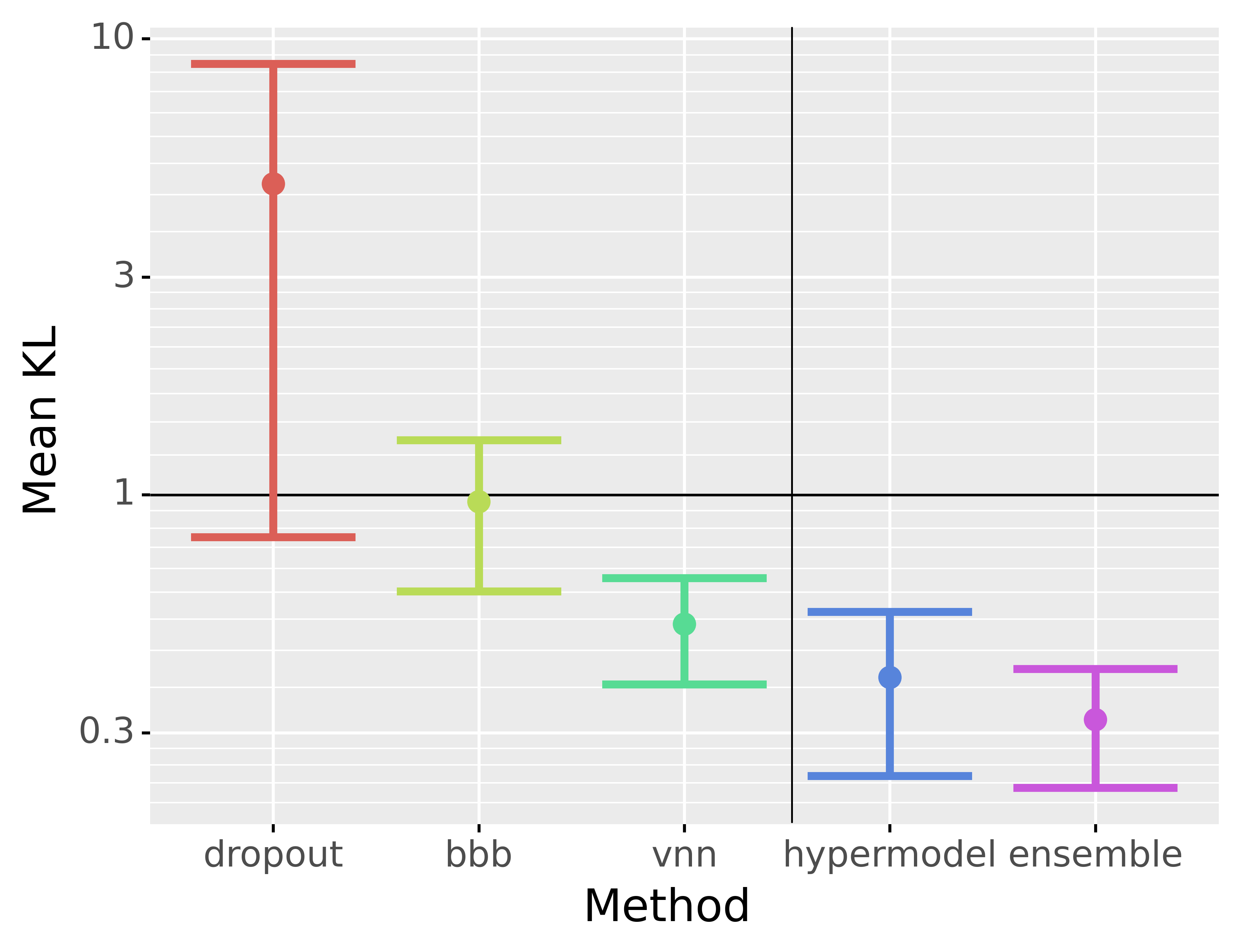}
         \caption{All parameters}
         \label{fig:results-kl-all}
     \end{subfigure}
     \begin{subfigure}[b]{1\linewidth}
         \centering
         \includegraphics[width=0.9\linewidth]{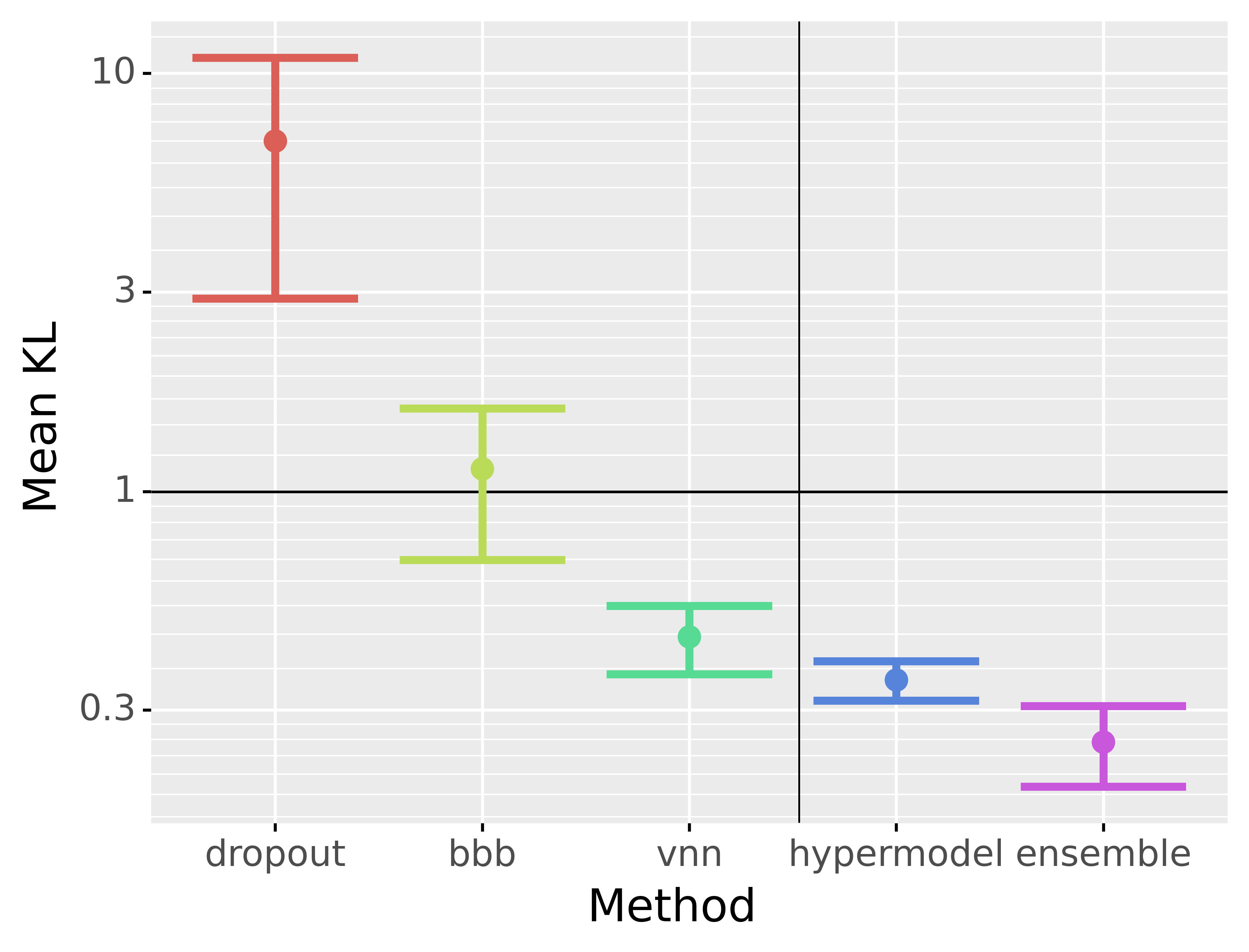}
         \caption{$D_x = 1000$}
         \label{fig:results-kl-1000}
     \end{subfigure}
    \caption{Comparison of mean KL value with 1 STD range for each method averaged across different experiment parameters.}
    \label{fig:results-kl}
\end{figure}

We implement VNNs inside the ENN's JAX implementation \cite{github_enn} to reproduce results for BBB \cite{blundell2015weight}, MCD \cite{2016dropout}, Ensemble \cite{osband2018randomized}, Hypermodel \cite{dwaracherla2020hypermodels} and compare them with VNN\footnote{The code is available at Anonymized for review process.}. 
We follow the original framework parameters and repeat experiments with the following options: $D_x \in \{10, 100, 1000\}$, $\lambda \in \{1, 10, 100\}$, and $\epsilon \in \{0.01, 0.1, 1\}$. 
Each model is trained with 10 different random seeds, and the resulting KL value is the average of individual runs.
The average KL values for all experiment parameters are given in Fig. \ref{fig:results-kl-all} and for the highest input dimension value $D_x=1000$ are given in Fig. \ref{fig:results-kl-1000}.
VNN has better uncertainty quality than BBB and MCD, but it is outperformed by Hypermodel and Ensemble. This can be explained by the difference in BMA for Deep Ensembles and Variation Inference methods, as explained in \cite{wilson2020bayesian}.
The weight probability distribution can be split into basins, where models sampled from the same basin are too similar and will describe the problem from the same point of view, resulting in multiple entries of actually identical model in the prediction voting.
Deep Ensembles and Hypermodels avoid this problem by not having a single anchor point with small weight deviations, and therefore having high chances of converging trained models into different basins. This means that VNN has a higher chance than Ensemble to have its samples in a single basin, placing it in the same group as BBB and MCD. 
Additionally, with bigger data dimensionality $D_x$, MCD and BBB achieve worse results, while VNN, Ensemble and Hypermodel perform better.

\begin{figure}
\centering
    \includegraphics[width=1\linewidth]{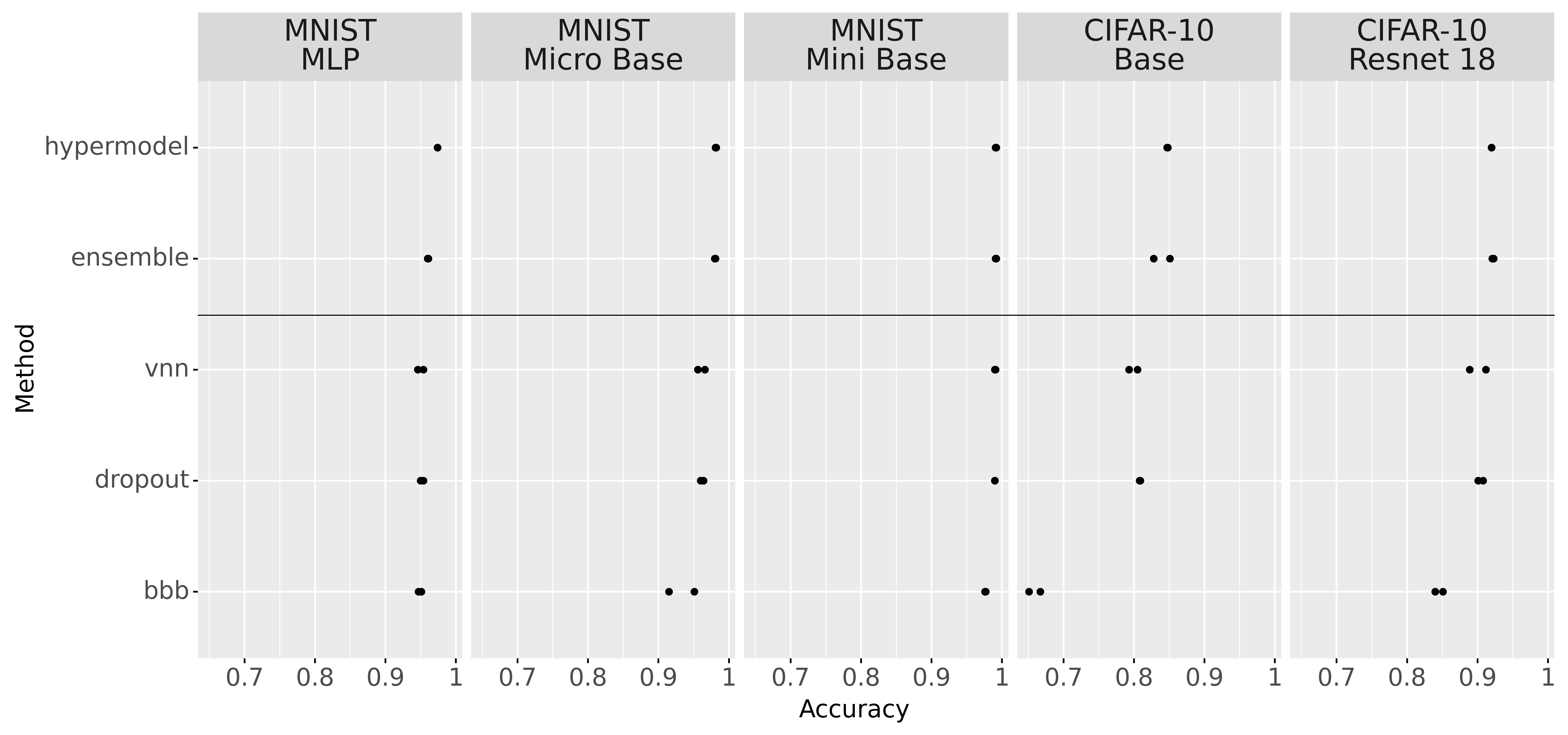}
    \caption{Comparison of classification accuracy on MNIST and CIFAR-10 datasets with different model architectures.}
    \label{fig:results-classification}
\end{figure}

We further perform experiments on image classification. We train the same methods for image classification tasks on MNIST \cite{deng2012mnist} and CIFAR-10 \cite{Krizhevsky09cifar10} datasets.
To show the influence of model architecture on the performance, we use a set of architectures $\{F_i\}$ and train each method with the selected architecture $F$.
We select a Base architecture to have 3 convolutional and 1 linear layer for MNIST, and 6 convolutional and 1 linear layer for CIFAR-10.
Mini and Micro Base architectures have the same layer structure as the Base one, but a lower number of channels in each layer.
MLP architecture consists of 3 fully connected layers.
We also use Resnet-18 \cite{he2015resnet} architecture for experiments on CIFAR-10. 
For each method, we train models with different hyperparameter values and select the best two models for comparison.
The results of classification experiments are given in Fig. \ref{fig:results-classification} and are roughly following the results of uncertainty quality estimation experiments.

\vspace{-0.3cm}
\section{Conclusion}\label{sec:conclusions}
We proposed Variational Neural Networks that consider a Gaussian distribution over outputs of each layer, the mean and variance of which are generated by the corresponding sub-layers, and evaluated their uncertainty estimation quality within the Epistemic Neural Networks framework.
Experiments show that, despite having similar properties of Bayesian Model Averaging to Monte Carlo Dropout and Bayes By Backpropagation, where sampled models are close resulting in similar models' points of view, VNNs achieve better uncertainty quality which is retained when data dimensionality is increased, in contrast to Monte Carlo Dropout and Bayes By Backpropagation methods.

\section*{Acknowledgment}
We thank Dr. Martin Magris for helpful discussions and feedback. 

\bibliographystyle{IEEEtran}
\bibliography{bibliography}

\begin{thebibliography}{10}
\providecommand{\url}[1]{#1}
\csname url@samestyle\endcsname
\providecommand{\newblock}{\relax}
\providecommand{\bibinfo}[2]{#2}
\providecommand{\BIBentrySTDinterwordspacing}{\spaceskip=0pt\relax}
\providecommand{\BIBentryALTinterwordstretchfactor}{4}
\providecommand{\BIBentryALTinterwordspacing}{\spaceskip=\fontdimen2\font plus
\BIBentryALTinterwordstretchfactor\fontdimen3\font minus
  \fontdimen4\font\relax}
\providecommand{\BIBforeignlanguage}[2]{{%
\expandafter\ifx\csname l@#1\endcsname\relax
\typeout{** WARNING: IEEEtran.bst: No hyphenation pattern has been}%
\typeout{** loaded for the language `#1'. Using the pattern for}%
\typeout{** the default language instead.}%
\else
\language=\csname l@#1\endcsname
\fi
#2}}
\providecommand{\BIBdecl}{\relax}
\BIBdecl

\bibitem{1995bayessian_nn}
D.~J.~C. Mackay, ``Probable networks and plausible predictions — a review of
  practical bayesian methods for supervised neural networks,'' \emph{Network},
  vol.~6, no.~3, pp. 469--505, 1995.

\bibitem{wilson2020bayesian}
A.~G. Wilson and P.~Izmailov, ``{Bayesian Deep Learning and a Probabilistic
  Perspective of Generalization},'' in \emph{{NeurIPS}}, vol.~33, 2020.

\bibitem{charnock2020bayesian}
T.~Charnock, L.~Perreault-Levasseur, and F.~Lanusse, ``Bayesian neural
  networks,'' \emph{arXiv:2006.01490}, 2020.

\bibitem{blundell2015weight}
C.~Blundell, J.~Cornebise, K.~Kavukcuoglu, and D.~Wierstra, ``{Weight
  Uncertainty in Neural Networks},'' in \emph{{ICML}}, vol.~37, 2015, pp.
  1613--1622.

\bibitem{2016dropout}
Y.~Gal and Z.~Ghahramani, ``Dropout as a bayesian approximation: Representing
  model uncertainty in deep learning,'' in \emph{{ICML}}, vol.~48, 2016, pp.
  1050--1059.

\bibitem{osband2018randomized}
I.~Osband, J.~Aslanides, and A.~Cassirer, ``Randomized prior functions for deep
  reinforcement learning,'' in \emph{{NeurIPS}}, vol.~31, 2018, pp. 8626--8638.

\bibitem{hastings1970mcmc}
W.~K. Hastings, ``{Monte Carlo sampling methods using Markov chains and their
  applications},'' \emph{Biometrika}, vol.~57, no.~1, pp. 97--109, 1970.

\bibitem{Blei2017vi}
D.~M. Blei, A.~Kucukelbir, and J.~D. McAuliffe, ``Variational inference: A
  review for statisticians,'' \emph{{JASA}}, vol. 112, no. 518, pp. 859--877,
  2017.

\bibitem{1951_kl_information}
S.~Kullback and R.~A. Leibler, ``{On Information and Sufficiency},'' \emph{Ann.
  Math. Stat.}, vol.~22, no.~1, pp. 79 -- 86, 1951.

\bibitem{hoffman2012svi}
M.~Hoffman, D.~M. Blei, C.~Wang, and J.~Paisley, ``Stochastic variational
  inference,'' \emph{{JMLR}}, vol.~14, no.~1, pp. 1303--1347, 2013.

\bibitem{magris2022bayes_survey}
M.~Magris and A.~Iosifidis, ``Bayesian learning for neural networks: an
  algorithmic survey,'' \emph{arxiv:2211.11865}, 2022.

\bibitem{kelley1960backprop}
H.~J. Kelley, ``Gradient theory of optimal flight paths,'' \emph{{ARSJ}},
  vol.~30, no.~10, pp. 947--954, 1960.

\bibitem{srivastava2014dropout}
N.~Srivastava, G.~Hinton, A.~Krizhevsky, I.~Sutskever, and R.~Salakhutdinov,
  ``Dropout: A simple way to prevent neural networks from overfitting,''
  \emph{{JMLR}}, vol.~15, no.~56, pp. 1929--1958, 2014.

\bibitem{dwaracherla2020hypermodels}
V.~Dwaracherla, X.~Lu, M.~Ibrahimi, I.~Osband, Z.~Wen, and B.~V. Roy,
  ``Hypermodels for exploration,'' in \emph{{ICLR}}, vol.~8, 2020.

\bibitem{kendal2017uncertainties}
A.~Kendall and Y.~Gal, ``What uncertainties do we need in bayesian deep
  learning for computer vision?'' in \emph{{NeurIPS}}, vol.~30, 2017.

\bibitem{kingma2014autoencoding}
D.~P. Kingma and M.~Welling, ``Auto-encoding variational bayes,'' in
  \emph{{ICLR}}, vol.~2, 2014.

\bibitem{osband2021epistemic}
I.~Osband, Z.~Wen, M.~Asghari, M.~Ibrahimi, X.~Lu, and B.~V. Roy, ``{Epistemic
  Neural Networks},'' \emph{arXiv:2107.08924}, 2021.

\bibitem{hullermeier2021epiale}
E.~H{\"{u}}llermeier and W.~Waegeman, ``Aleatoric and epistemic uncertainty in
  machine learning: an introduction to concepts and methods,'' \emph{Machine
  Learning}, vol. 110, no.~3, pp. 457--506, 2021.

\bibitem{gawlikowski2021uncertaintyindl}
J.~Gawlikowski, C.~R.~N. Tassi, M.~Ali, J.~Lee, M.~Humt, J.~Feng, A.~M. Kruspe,
  R.~Triebel, P.~Jung, R.~Roscher, M.~Shahzad, W.~Yang, R.~Bamler, and X.~X.
  Zhu, ``A survey of uncertainty in deep neural networks,''
  \emph{arxiv:2107.03342}, 2021.

\bibitem{allenby2006hierarchicalbayes}
G.~M. Allenby and P.~E. Rossi, ``Hierarchical bayes models,'' \emph{The
  handbook of marketing research}, pp. 418--440, 2006.

\bibitem{lee2018nngp}
J.~Lee, Y.~Bahri, R.~Novak, S.~S. Schoenholz, J.~Pennington, and
  J.~Sohl-Dickstein, ``Deep neural networks as gaussian processes,'' in
  \emph{{ICLR}}, 2018.

\bibitem{github_enn}
I.~Osband, Z.~Wen, M.~Asghari, M.~Ibrahimi, X.~Lu, and B.~V. Roy, ``Github code
  for epistemic neural networks,'' \url{https://github.com/deepmind/enn}, 2021.

\bibitem{deng2012mnist}
L.~Deng, ``The mnist database of handwritten digit images for machine learning
  research,'' \emph{IEEE Signal Process. Mag.}, vol.~29, no.~6, pp. 141--142,
  2012.

\bibitem{Krizhevsky09cifar10}
A.~Krizhevsky, ``Learning multiple layers of features from tiny images,'' Tech.
  Rep., 2009.

\bibitem{he2015resnet}
K.~He, X.~Zhang, S.~Ren, and J.~Sun, ``Deep residual learning for image
  recognition,'' in \emph{{CVPR}}, 2015, pp. 770--778.

\end{thebibliography}

\end{document}